\documentclass[11pt,a4paper]{article}
\usepackage[hyperref]{acl2021}
\usepackage{times}
\usepackage{latexsym}
\usepackage{booktabs}
\usepackage{multirow}
\usepackage{todonotes}
\usepackage{stfloats} 

\usepackage{microtype}

\usepackage{graphicx}

\aclfinalcopy 

\title{FLERT: Document-Level Features for Named Entity Recognition}

\author{
	Stefan Schweter \\
	schweter.ml\\
	{\tt stefan@schweter.eu}\\\And
	Alan Akbik \\
	Humboldt-Universität zu Berlin\\
	{\tt alan.akbik@hu-berlin.de}\\
\\}

\date{}

\begin{document}
\maketitle
\begin{abstract}

Current state-of-the-art approaches for named entity recognition (NER) typically consider text at the sentence-level and thus do not model information that crosses sentence boundaries. However, the use of transformer-based models for NER offers natural options for capturing document-level features. In this paper, we perform a comparative evaluation of document-level features in the two standard NER architectures commonly considered in the literature, namely "fine-tuning" and "feature-based LSTM-CRF". We evaluate different hyperparameters for document-level features such as context window size and enforcing document-locality. We present experiments from which we derive recommendations for how to model document context and present new state-of-the-art scores on several CoNLL-03 benchmark datasets. Our approach is integrated into the \textsc{Flair} framework to facilitate reproduction of our experiments. 

\end{abstract}

\section{Introduction}
\vspace{-1mm}

Named entity recognition (NER) is the well-studied NLP task of predicting shallow semantic labels for sequences of words, used for instance for identifying the names of persons, locations and organizations in text. Current approaches for NER often leverage pre-trained transformer architectures such as BERT \citep{devlin-etal-2019-bert} or XLM \citep{lample2019cross}.

\begin{figure*}
\vspace{-3mm}
 \centering
 \includegraphics[width=\linewidth]{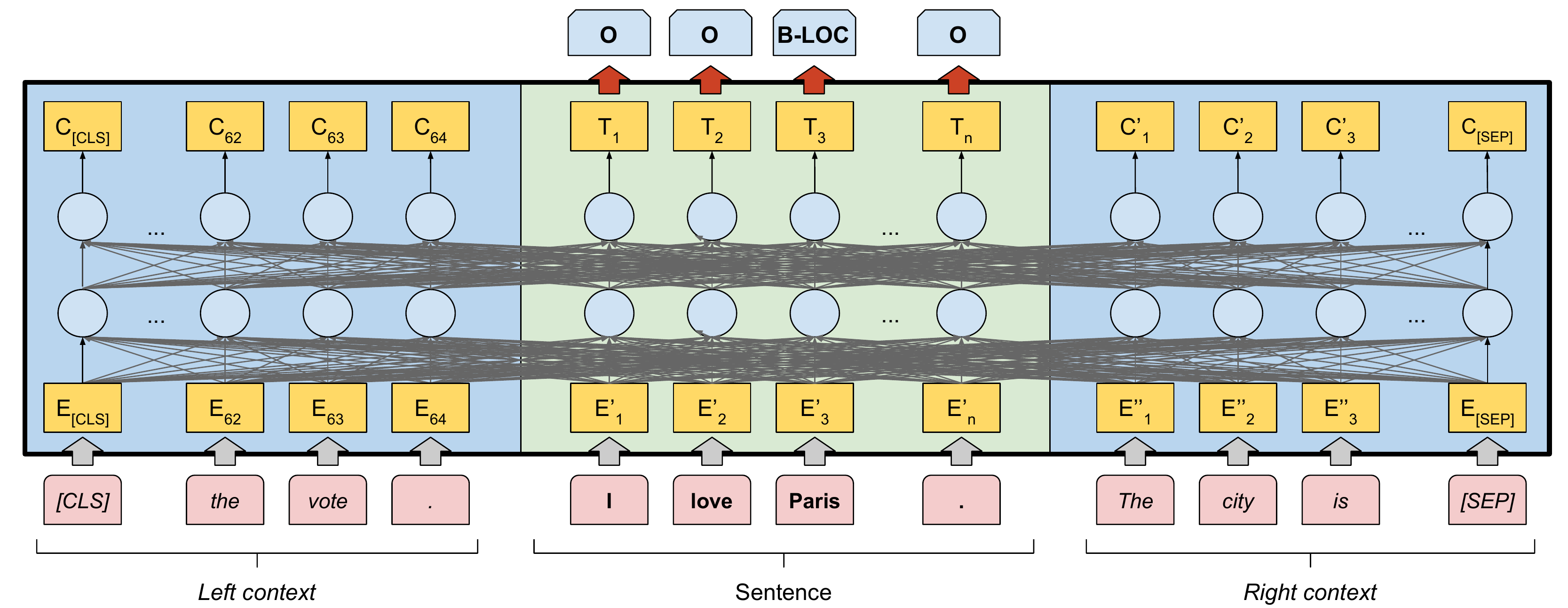}
\vspace{-7mm}
 \caption{To obtain document-level features for a sentence that we wish to tag ("I love Paris", shaded green), we add 64 tokens of left and right tokens each (shaded blue). As self-attention is calculated over all input tokens, the representations for the sentence's tokens are influenced by the left and right context.}
\vspace{-3mm}
 \label{overview-bert-fine-tuning-document-level}
\end{figure*}

\noindent 
\textbf{Document-level features.} While NER is traditionally modeled at the sentence-level, transformer-based models offer a natural option for capturing document-level features by passing a sentence with its surrounding context. As Figure~\ref{overview-bert-fine-tuning-document-level} shows, this context can then influence the word representations of a sentence: The example sentence "I love Paris" is passed through the transformer together with the next sentence that begins with "The city is", potentially helping to resolve the ambiguity of the word "Paris".  
A number of prior works have employed such document-level features~\cite{devlin-etal-2019-bert, virtanen2019multilingual,yu-etal-2020-named} but only in combination with other contributions and thus have not evaluated the impact of using document-level features in isolation. 

\noindent
\textbf{Contributions.}
With this paper, we close this experimental gap and present an evaluation of document-level features for NER. As there are two conceptually very different approaches for transformer-based NER that are currently used across the literature, we evaluate document-level features in both: 

\begin{enumerate}
\vspace{-1mm}
    \item In the first, we \textit{fine-tune} the transformer itself on the NER task and only add a linear layer for word-level predictions~\cite{devlin-etal-2019-bert}.
\vspace{-1mm}
    \item In the second, we use the transformer only to provide \textit{features} to a standard LSTM-CRF sequence labeling architecture~\cite{2015arXiv150801991H} and thus perform no fine-tuning.
\end{enumerate}
\vspace{-1mm}
We discuss the differences between both approaches and explore best hyperparameters for each. In their best determined setup, we then perform a comparative evaluation.
We find that (1) document-level features significantly improve NER quality and that (2) fine-tuning generally outperforms feature-based approaches. We also determine best settings for document-level context and report several new state-of-the-art scores on the classic CoNLL benchmark datasets. Our approach is integrated as the "FLERT"-extension into the \textsc{Flair} framework~\cite{akbik-etal-2019-flair} to facilitate further experimentation.

\section{Document-Level Features}
\vspace{-2mm}

In a transformer-based architecture, document-level features can easily be realized by passing a sentence with its surrounding context to obtain word embeddings, as illustrated in Figure~\ref{overview-bert-fine-tuning-document-level}. 

\noindent
\textbf{Prior approaches.}
This approach was first employed by~\citet{devlin-etal-2019-bert} with what they described as a "maximal document context", though technical details were not listed. Subsequent work used variants of this approach. For instance, \citet{virtanen2019multilingual} experiment with adding the following (but not preceding) sentence as context to each sentence. \citet{yu-etal-2020-named} instead use a 64 surrounding token window for each token in a sentence, thus calculating a large context on a per-token basis. By contrast, \citet{luoma2020exploring} adopt a multi-sentence view in which they combine predictions from different windows and sentence positions.  

\noindent
\textbf{Our approach.}
In this paper, we instead use a conceptually simple variant in which we create context on a per-sentence basis: For each sentence we wish to classify, we add 64 subtokens of left and right context, as shown in Figure~\ref{overview-bert-fine-tuning-document-level}. This has computational and implementation advantages in that each sentence and its context need only be passed through the transformer once and that added context is limited to a relatively small window. Furthermore, we can still follow standard procedure in shuffling  sentences at each epoch during training, since context is encoded on a per-sentence level. We use this approach throughout this paper.

\section{Baseline Parameter Experiments}
\label{sec:baseline_experiments}
\vspace{-2mm}

As mentioned in the introduction, there are two common architectures for transformer-based NER, namely fine-tuning and feature-based approaches. In this section, we briefly introduce the differences between both approaches and conduct a study to identify best hyperparameters for each. The best respective setups are then used in the final comparative evaluation in Section~\ref{sec:evaluation}. 



\begin{table*}
\vspace{-2mm}
\centering
\resizebox{440px}{!}{
\begin{tabular}{lccccc}
\hline
 Fine-tuning Approach &  \textsc{En} & \textsc{De}      & $\textsc{De}_{06}$ & \textsc{Nl} & \textsc{Es} \\ \hline
Transformer-Linear       &      96.64 $\pm$ 0.14        & 89.06 $\pm$ 0.18 & 91.86 $\pm$ 0.41   &    93.41 $\pm$ 0.19         & 88.95 $\pm$ 0.19 \\
 + \textit{Document features}            &   96.82 $\pm$ 0.07           & \textbf{89.79} $\pm$ 0.13 & \textbf{93.09} $\pm$ 0.06   &     94.19 $\pm$ 0.14        & 90.34 $\pm$ 0.27 \\ 
  + WE                 &     96.82 $\pm$ 0.13         & 88.96 $\pm$ 0.10 & 92.12 $\pm$ 0.10   &    93.51 $\pm$ 0.09         & 89.09 $\pm$ 0.36 \\
+ WE  + \textit{Document features}       &    \textbf{97.02 $\pm$ 0.09}          & 89.74 $\pm$ 0.46 & 92.83 $\pm$ 0.12   &   94.01 $\pm$ 0.27          & 90.17 $\pm$ 0.25 \\ \hline
 Transformer-CRF   &    96.79 $\pm$ 0.11          & 88.52 $\pm$ 0.10 & 92.21 $\pm$ 0.07   &    93.61 $\pm$ 0.15         & 88.77 $\pm$ 0.20 \\ 
 + \textit{Document features}            &    96.90 $\pm$ 0.06          & 89.67 $\pm$ 0.24 & 92.87 $\pm$ 0.21   &     94.16 $\pm$ 0.07        & \textbf{90.56} $\pm$ 0.09 \\ 
  + WE                 &   96.79 $\pm$ 0.15           & 88.84 $\pm$ 0.15 & 91.97 $\pm$ 0.09   &    93.36 $\pm$ 0.04         & 88.63 $\pm$ 0.47 \\

+ WE + \textit{Document features}        &   96.87 $\pm$ 0.00           & 89.69 $\pm$ 0.22 & 92.88 $\pm$ 0.26   &     \textbf{94.34 $\pm$ 0.13}        & 90.37 $\pm$ 0.14 \\\hline
\end{tabular}%
}
\vspace{-1mm}
\caption{Evaluation of different variants using the fine-tuning approach. The evaluation is performed against the \textbf{development set} of all 4 languages of the CoNLL-03 shared task for NER.}
\label{tab:finetuning_development_dataset}
\vspace{-3mm}
\end{table*}
\subsection{Setup}

\noindent
\textbf{Data set.} We use the development datasets of the CoNLL shared tasks \citep{tjong-kim-sang-de-meulder-2003-introduction,tjong-kim-sang-2002-introduction} for NER on four languages (English, German, Dutch and Spanish). Following \citet{yu-etal-2020-named} we report results for both the original and revised dataset for German (denoted as $\textsc{De}_{06}$).

\noindent
\textbf{Transformer model.}
In all experiments in this section, we employ the multilingual XLM-RoBERTa (XLM-R) transformer model proposed by \citet{conneau2019unsupervised}. We use the \texttt{xlm-roberta-large} model in our experiments, trained on 2.5TB of data from a cleaned Common Crawl corpus \cite{wenzek-etal-2020-ccnet} for 100 different languages


\noindent
\textbf{Embeddings (+WE).}
For each setup we experiment with concatenating classic word embeddings to the word-level representations obtained from the transformer model. Following~\citet{akbik-etal-2018-contextual}, we use \textsc{GloVe} embeddings \citep{pennington-etal-2014-glove} for English and \textsc{FastText} embeddings \citep{bojanowski-etal-2017-enriching} for other languages.


\subsection{First Approach: Fine-Tuning}
\label{sec:fine_tuning}
\vspace{-1mm}

Fine-tuning approaches typically only add a single linear layer to a transformer and fine-tune the entire architecture on the NER task. To bridge the difference between subtoken modeling and token-level predictions, they apply \textit{subword pooling} to create token-level representations which are then passed to the final linear layer. Conceptually, this approach has the advantage that everything is modeled in a single architecture that is fine-tuned as a whole. More details on parameters and architecture are provided in the Appendix. 


\noindent 
\textbf{Evaluated variants.} 
We compare two variants: 

\begin{description}
\vspace{-1mm}
\item[Transformer-Linear] In the first, we use the standard approach of adding a simple linear classifier on top of the transformer to directly predict tags. 
\vspace{-1mm}
\item[Transformer-CRF] In the second, we evaluate if it is helpful to add a conditional random fields (CRF) decoder between the transformer and the linear classifier~\cite{souza2019portuguese}. 
\end{description}
\vspace{-1mm}
\noindent
Results are listed in Table~\ref{tab:finetuning_development_dataset}.

\subsection{Second Approach: Feature-Based}
\label{sec:feature_based}
\vspace{-1mm}

Feature-based approaches instead use the transformer only to generate embeddings for each word in a sentence and use these as input into a standard sequence labeling architecture, most commonly a LSTM-CRF~\cite{2015arXiv150801991H}. The transformer weights are frozen so that training is limited to the LSTM-CRF. Conceptually, this approach benefits from a well-understood model training procedure that includes a real stopping criterion. See Appendix B for more details on training parameters. 




\noindent 
\textbf{Evaluated variants.} 
We compare two variants:
\begin{description}
\vspace{-1mm}
\item[All-layer-mean] In the first, we obtain embeddings for each token using mean pooling across all transformer layers, including the word embedding layer. This representation has the same length as the hidden size for each transformer layer. This approach is inspired by the ELMO-style~\cite{peters-etal-2018-deep} "scalar mix".
\vspace{-1mm}
\item[Last-four-layers] In the second, we follow~\citet{devlin-etal-2019-bert} to only use the last four transformer layers for each token and concatenate their representations into a final representation for each token. It thus has four times the length of the transformer layer hidden size.
\end{description}
The results for English\footnote{Other languages show similar results (omitted for space).} are shown in Table~\ref{tab:feature_based_development_dataset}.
\begin{table}
\centering
\resizebox{190px}{!}{
\begin{tabular}{lc}
\hline
 Feature-based Approach             & \textsc{En}                \\ \hline 
 \textsc{LSTM-CRF} (last-four-layers)  & 91.17 $\pm$ 0.29           \\        
 + \textit{Document features}                          & 94.23 $\pm$ 0.19           \\        
  + WE                               & 92.19 $\pm$ 0.46           \\        

  + WE   + \textit{Document features}                  & 94.61 $\pm$ 0.10           \\ \hline 
 \textsc{LSTM-CRF} (all-layer-mean)    & 94.37 $\pm$ 0.06           \\        
 + \textit{Document features}                          & 96.09 $\pm$ 0.07           \\        
  + WE                               & 95.63 $\pm$ 0.04           \\        

 + WE + \textit{Document features}                      & \textbf{96.53} $\pm$ 0.10  \\ \hline 
\end{tabular}%
}
\vspace{-2mm}
\caption{Evaluation of feature-based approach on CoNLL-03 \textbf{development set}.}
\label{tab:feature_based_development_dataset}
\vspace{-4mm}
\end{table}

\begin{table*}
\vspace{-2mm}
\centering
\resizebox{\textwidth}{!}{
\begin{tabular}{lllllll}
\hline
 Approach &  \textit{Doc. features?} & \textsc{En} & \textsc{De} & $\textsc{De}_{06}$ & \textsc{Nl} & \textsc{Es} \\ \hline
 \textit{Feature-based}  & & & & & & \\ 
 
   LSTM-CRF (all layer mean) & no & 91.83 $\pm$ 0.06 & 82.88 $\pm$ 0.28 & 87.35 $\pm$ 0.17 & 89.87 $\pm$ 0.45 & 88.78 $\pm$ 0.08\\ 
    LSTM-CRF (all layer mean) & yes & 93.12 $\pm$ 0.14 & 84.86 $\pm$ 0.11 & 89.88 $\pm$ 0.26 & 91.73 $\pm$ 0.21 & 88.98 $\pm$ 0.11\\  \hline
 \textit{Fine-tuning}  & & & & & & \\ 
   Transformer-Linear  & no & 92.79 $\pm$ 0.10 & 86.60 $\pm$ 0.43 & 90.04 $\pm$ 0.37 & 93.50 $\pm$ 0.15 & 89.94 $\pm$ 0.24 \\ 
  Transformer-Linear & yes &  93.64 $\pm$ 0.05 & 86.99 $\pm$ 0.24 & 91.55 $\pm$ 0.07 & 94.87 $\pm$ 0.20 & 90.14 $\pm$ 0.14 \\ \hline
  \textit{Fine-tuning (Ablations)}  & & & & & & \\ 
  
   Transformer-Linear &  yes (\textit{+enforce})  & 93.75 $\pm$ 0.16 & 87.35 $\pm$ 0.15 & 91.33 $\pm$ 0.18 & \textbf{95.21 $\pm$ 0.08} & -- \\ 
  
  
   Transformer-Linear (\textsc{+dev}) &   yes (\textit{+enforce})     & 94.09 $\pm$ 0.07 & \textbf{88.34 $\pm$ 0.36} & \textbf{92.23 $\pm$ 0.21} & 95.19 $\pm$ 0.32  & -- \\  \hline

 
 \textit{Best published} & & & & & & \\ 
 \citet{akbik2019naacl} & pooling & 93.18 $\pm$ 0.09 & -- & 88.27 $\pm$ 0.30 & 90.44 $\pm$ 0.20 & -- \\
 \citet{yu-etal-2020-named} & yes & 93.5 & 86.4 & 90.3 & 93.7 & \textbf{90.3} \\
 \citet{strakova-etal-2019-neural} & yes & 93.38 & 85.10 & -- & 92.69 & 88.81 \\ 
 \citet{yamada-etal-2020-luke} & yes & \textbf{94.3} & -- & -- & -- & -- \\\hline
 
\end{tabular}%
}
\vspace{-2mm}
\caption{Comparative evaluation of best configurations of fine-tuning and feature-based approaches on test data.
}
\label{tab:stats_best_configurations}
\vspace{-2mm}
\end{table*}





\subsection{Results: Best Configurations}
\vspace{-1mm}

We evaluate both approaches in each variant in all possible combinations of adding standard word embeddings "(+WE)" and document-level features "(+\textit{Document features})". Each setup is run three times to report average F1 and standard deviation.

\noindent
\textbf{Results.} For fine-tuning, we find that additional word embeddings and using a CRF decoder improves results only for some languages, and often only minimally so (see Table~\ref{tab:finetuning_development_dataset}). We thus choose a minimal Transformer-Linear architecture. For the feature-based approach, we find that an all-layer-mean strategy and adding word embeddings very clearly yields the best results (see Table~\ref{tab:feature_based_development_dataset}).



\begin{table}
\centering
\resizebox{220pt}{!}{
\begin{tabular}{ccccccc}
\hline
 CW &  \textsc{En}   & \textsc{De}    & $\textsc{De}_{06}$ & \textsc{Nl}    & \textsc{Es}    & Avg. \\ \hline
 48             & 96.86			 & 89.47		  & 92.63			   & 94.09		    & 90.31	         & 92.67 \\
 64             & 96.82          & 89.64          & \textbf{92.87}     & 94.19          & \textbf{90.34} & \textbf{92.77} \\
 96             & \textbf{96.90} & \textbf{89.67} & 92.58              & 94.03          & 90.31          & 92.70 \\
 128            & \textbf{96.90} & 88.97          & 92.56              & \textbf{94.22} & 90.15          & 92.56 \\
 \hline
\end{tabular}%
}
\vspace{-2mm}
\caption{Comparative evaluation of context window sizes of fine-tuning approach on development set.}
\label{tab:comparison_different_context_window_sizes}
\end{table}

\section{Comparative Evaluation}
\label{sec:evaluation}
\vspace{-1mm}

With the best configurations identified in Section~\ref{sec:baseline_experiments} on the development data,    
we conduct a final comparative evaluation on the test splits of the CoNLL-03 datasets, with and without document features.




\subsection{Main Results}

\vspace{-1mm}
The evaluation results are listed in Table~\ref{tab:stats_best_configurations}. We make the following observations:
 

\begin{table}
\centering
\resizebox{220pt}{!}{
\begin{tabular}{lccccc}
\hline
 Entity &  \textsc{En} & \textsc{De}       & $\textsc{De}_{06}$ & \textsc{Nl}       & \textsc{Es} \\ \hline
 {\tt LOC}    & +0.44        & +0.23             & +1.97              & \underline{-0.74} & +0.17 \\
 {\tt MISC}   & +0.22        & \underline{-0.90} & +1.66              & +1.16             & +0.72 \\
 {\tt ORG}    & +1.21        & +0.56             & +0.74              & +1.66             & +0.11 \\
 {\tt PER}    & +1.19        & +1.15             & +1.50              & \underline{-0.34} & +0.14 \\
 \hline
\end{tabular}%
}
\vspace{-2mm}
\caption{Relative change in F1 for different entity types and languages when adding document-level features.}
\label{tab:comparison_context_vs_no_context}
\vspace{-2mm}
\end{table}

\noindent
\textbf{Fine-tuning document-level features best.} As Table~\ref{tab:stats_best_configurations} shows, we find that fine-tuning outperforms the feature-based approach across all experiments ($\approx \uparrow$2 pp on average). 
Similarly, we find that document-level features clearly outperform sentence-level features ($\uparrow$1.15 pp on average). We thus find fine-tuning with document-level features to work best across all languages.

\noindent
\textbf{Enforcing document boundaries.} For fine-tuning, we also test an ablation in which we truncate document-features at document boundaries, meaning that context can only come from the same document. As the columns "yes (\textit{+enforce})" in Table~\ref{tab:stats_best_configurations} show, this increases F1-score across nearly all experiments. Our initial expectation that transformers would learn automatically to respect document boundaries (marked up in all datasets except Spanish) did not materialize, thus we recommend enforcing document boundaries if possible.

\noindent 
\textbf{New state-of-the-art results.} 
Combining fine-tuning, and strict document-level features yields new state-of-the-art scores for several datasets. Especially when including dev data in training (indicated as \textsc{+dev} in Table~\ref{tab:stats_best_configurations}) as is possible for fine-tuning as no stopping criterion is used. For German, we outperform ~\cite{yu-etal-2020-named} by $\uparrow$1.81 pp and $\approx \uparrow$2 pp on the original and revised German datasets respectively. For Dutch, we see an increase of by $\uparrow$1.5 pp over the next best approach. While we do not set new state-of-the-art scores for English and Spanish, our results are very competitive.

\subsection{Analysis}
\vspace{-1mm}
\noindent
\textbf{Impact of context window size (Table~\ref{tab:comparison_different_context_window_sizes}).}
We evaluate the impact of the number of surrounding tokens used in document-level contexts on
performance using the best configuration for fine-tuning approach. The context window is searched in $[48, 64, 96, 128]$. As Table~\ref{tab:comparison_different_context_window_sizes} shows, impact is marginal, with 64 the best across languages. 

\noindent
\textbf{Entity type analysis (Table \ref{tab:comparison_context_vs_no_context}).}
We perform a per-type analysis to compare average results across entity types with and without document-level features. We find that while the difference in F1-score depend on the type and the language, in particular the 
{\tt ORG} (organization) and {\tt PER} (person) entity types improve the most when including document-level features, indicating that cross-sentence contexts are most important here.





\vspace{-1mm}
\section{Conclusion}
\vspace{-1mm}

We evaluated document-level features in two commonly used NER architectures, for which we determined best setups. Our experiments show that document-level features significantly improve overall F1-score and that fine-tuning outperforms the feature-based LSTM-CRF. We also surprisingly find that enforcing document boundaries improves results, potentially adding to recent evidence that transformers have difficulties in learning positional signals~\cite{huang-etal-2020-improve}. We integrate our approach as the "FLERT"-extension\footnote{To be released with \textsc{Flair} version 0.8.} into the \textsc{Flair} framework, to enable the research community to leverage our best determined setups for training and applying state-of-the-art NER models.

\bibliography{anthology,acl2021}
\bibliographystyle{acl_natbib}
\newpage
\appendix
\clearpage
\newpage
\section{Appendix}

\subsection{Training: Fine-tuning Approach}

\begin{figure}[t!]
 \centering
 \includegraphics[width=\linewidth]{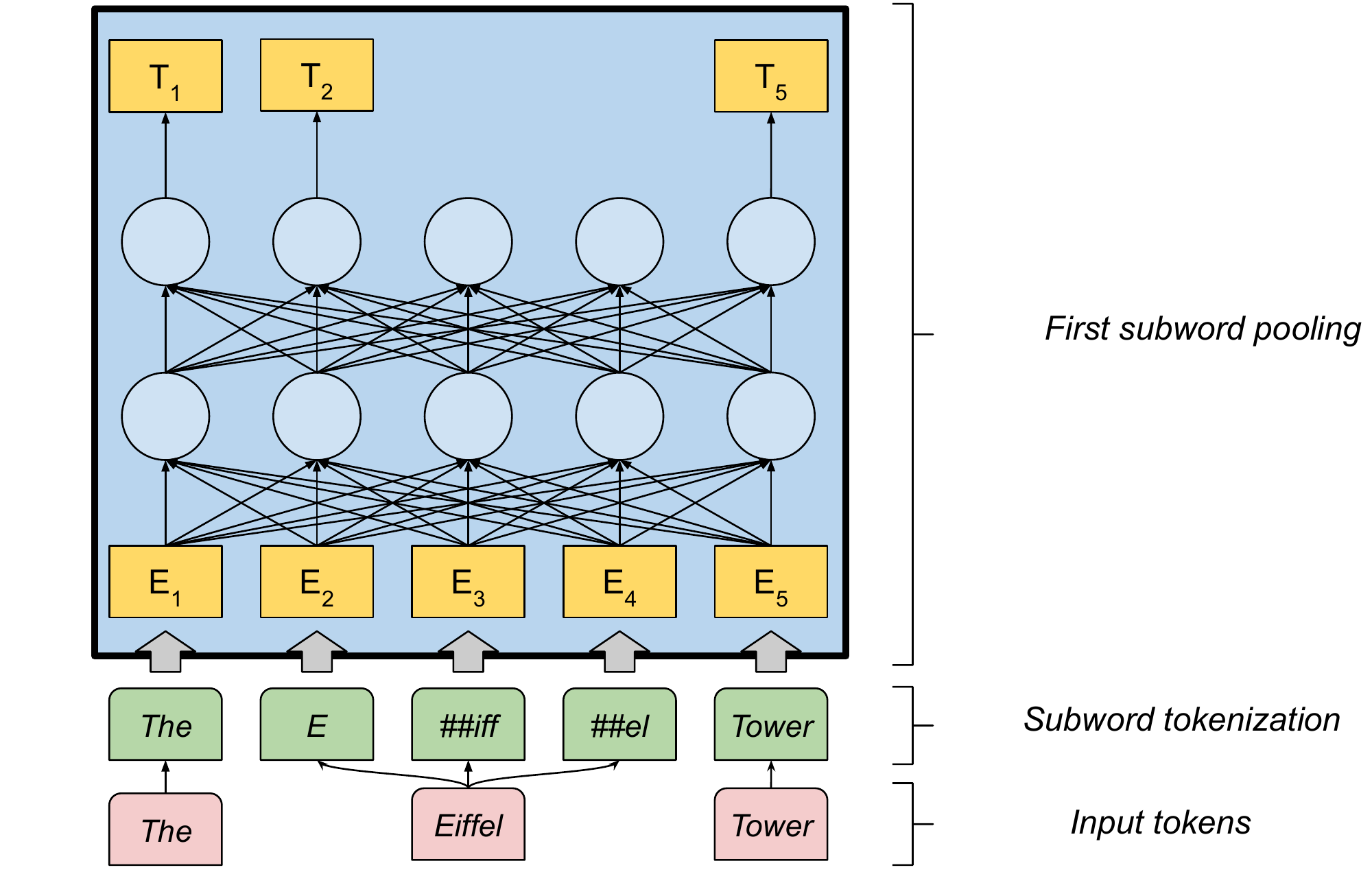}
 \captionof{figure}{
 Illustration of first subword pooling. The input "The Eiffel Tower" is subword-tokenized, splitting "Eiffel" into three subwords (shaded green). Only the first ("E") is used as representation for "Eiffel".
}
 \label{fig:subword_pooling}
\end{figure}

Fine-tuning only adds a single linear layer to a transformer and fine-tunes the entire architecture on the NER task. To bridge the difference between subtoken modeling and token-level predictions, they apply \textit{subword pooling} to create token-level representations which are then passed to the final linear layer. A common subword pooling strategy is "first"~\cite{devlin-etal-2019-bert} which uses the representation of the first subtoken for the entire token. See Figure~\ref{fig:subword_pooling} for an illustration.

\begin{table}[h]
\begin{center}
\begin{tabular}{ l l }
\toprule
Parameter & Value \\
\midrule
Transformer layers & last \\
Learning rate & 5e-6 \\
Mini batch size & 4 \\
Max epochs & 20 \\
Optimizer & AdamW \\ 
Scheduler & One-cycle LR \\
Subword pooling & first \\
\bottomrule
\end{tabular}
\end{center}
\vspace{-2mm}
\caption{\label{tab:finetuning_params} Parameters used for fine-tuning.}
\end{table}

\noindent 
\textbf{Training procedure.} 
To train this architecture, prior works typically use the AdamW~\cite{loshchilov2018decoupled} optimizer, a very small learning rate and a small, fixed number of epochs as a hard-coded stopping criterion~\cite{conneau2019unsupervised}. We adopt a one-cycle training strategy~\cite{2018arXiv180309820S}, inspired from the HuggingFace transformers~\cite{wolf2019huggingface} implementation, in which the learning rate linearly decreases until it reaches $0$ by the end of the training. Table~\ref{tab:finetuning_params} lists the architecture parameters we use across all our experiments.

\subsection{Training: Feature-based Approach}

\begin{figure*}[t]
 \centering
 \includegraphics[width=\linewidth]{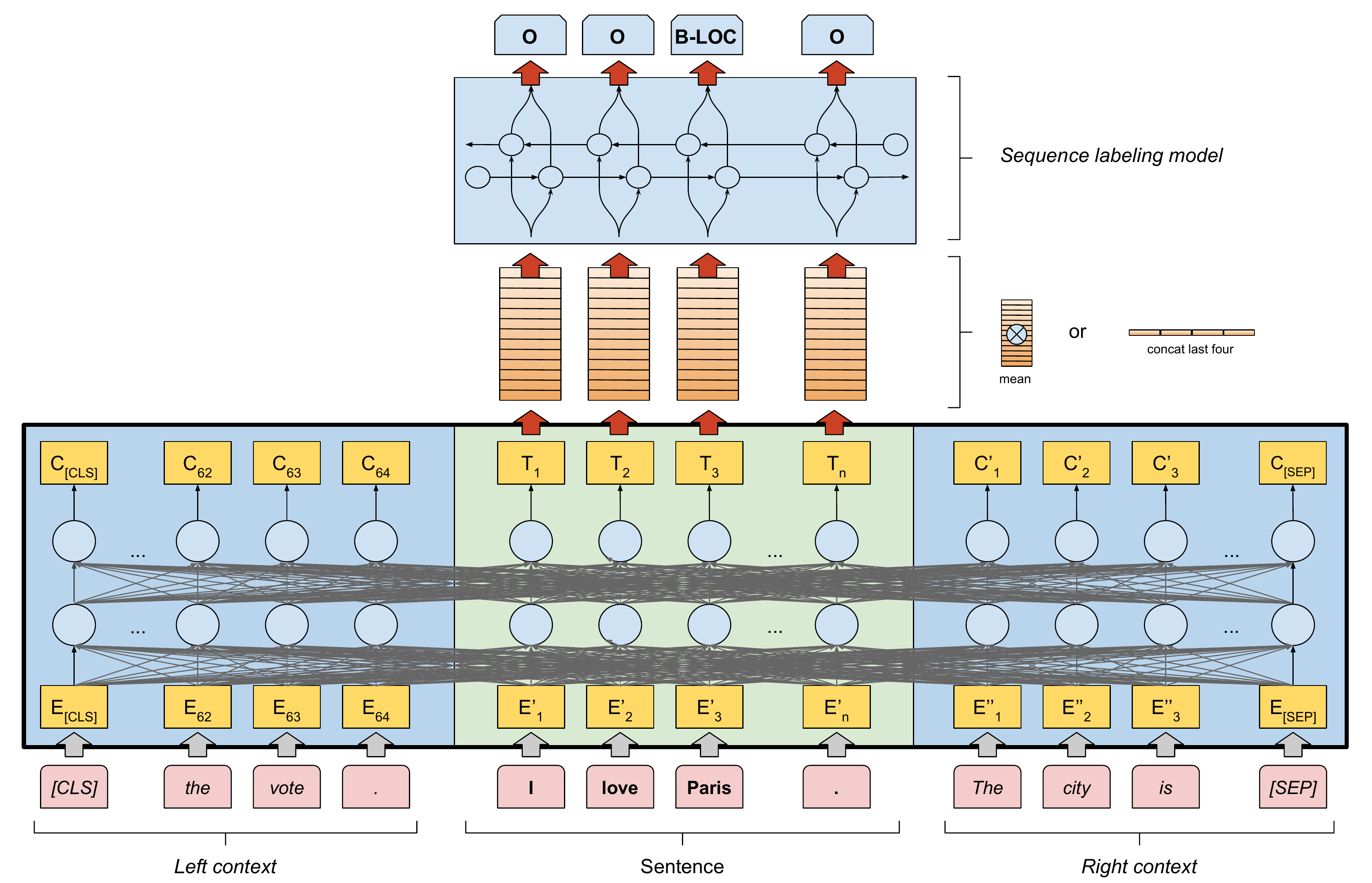}
 \captionof{figure}{
 Overview of feature-based approach. Self-attention is calculated over all input tokens (incl. left and right context). The final representation for each token in the sentence ("I love Paris", shaded green) can be calculated as a) mean over all layers of transformer-based model or b) concatenating the last four layers. 
}
 \label{overview-bert-feature-based-document-level}
\end{figure*}

Figure~\ref{overview-bert-feature-based-document-level} gives an overview of the feature-based approach: Word representations are extracted from the transformer by either averaging over all layers (all-layer-mean) or by concatenating the representations of the last four layers (last-four-layers). These are then input into a standard LSTM-CRF architecture~\cite{2015arXiv150801991H} as features. We again use the subword pooling strategy illustrated in Figure~\ref{fig:subword_pooling}. 


\noindent 
\textbf{Training procedure.} We adopt the standard training procedure used in earlier works. We with SGD with a larger learning rate that is annealed against the development data. Training terminates when the learning rate becomes too small. 
The parameters used for training a feature-based model are shown in Table~\ref{ner-training-parameters-feature-based}.

\begin{table}[h!]
\begin{center}
\begin{tabular}{ l l }
\toprule
Parameter & Value \\
\midrule
LSTM hidden size & 256 \\
Learning rate & 0.1 \\
Mini batch size & 16 \\
Max epochs & 500 \\
Optimizer & SGD \\
Subword pooling & first \\
\bottomrule
\end{tabular}
\end{center}
\caption{\label{ner-training-parameters-feature-based} Parameters for feature-based approach.}
\end{table}

\subsection{Reproducibility Checklist}

\noindent 
\textbf{Dataset statistics.} Table \ref{tab:dataset_stats} shows the number of sentences for for each dataset.

\begin{table}[h!]
\centering
\begin{tabular}{lllll}
\hline
 Split       &  \textsc{En} & \textsc{De} / $\textsc{De}_{06}$ & \textsc{Nl} & \textsc{Es} \\ \hline
 Train       & 14,987       & 12,705                           & 16,093      & 8,323 \\
 Dev         & 3,466        & 3,068                            & 2,969       & 1,915 \\
 Test        & 3,684        & 3,160                            & 5,314       & 1,517 \\
 \hline
\end{tabular}%
\caption{Number of sentences for each CoNLL dataset.}
\label{tab:dataset_stats}
\end{table}

\noindent
\textbf{Average training runtime.} We conduct experiments on a NVIDIA V-100 (16GB) for fine-tuning and a NVIDIA RTX 3090 TI (24GB) for the feature-based approach. We report average training times for our best configurations in Table \ref{tab:training_runtimes}.

\begin{table}[h!]
\centering
\begin{tabular}{ccccc}
\hline
 Approach       &  \textsc{En} & \textsc{De} / $\textsc{De}_{06}$ & \textsc{Nl} & \textsc{Es} \\ \hline
 Fine-Tuning    &  10h            & 10h                        & 10h             &  5h \\
 Feature-based  &  7h          &  5.5h                 &  5.75h      & 5.5h \\
 \hline
\end{tabular}%
\caption{Average training runtimes for our approaches.}
\label{tab:training_runtimes}
\end{table}

\noindent
\textbf{Number of model parameters.} The reported number of model parameters from \citet{conneau2019unsupervised} for XLM-R is 550M. Our fine-tuned model has 560M parameters ($\uparrow$1.8\%), whereas the feature-based model comes with 564M parameters ($\uparrow$2.5\%).

\noindent
\textbf{Evaluation metrics.} We evaluate our models using the CoNLL-2003 evaluation script\footnote{\url{https://www.clips.uantwerpen.be/conll2003/ner/bin/conlleval}} and report averaged F1-score over three runs.

\end{document}